# End-to-end Deep Learning Methods for Automated Damage Detection in Extreme Events at Various Scales

Yongsheng Bai
Department of Civil,
Environmental and Geodetic
Engineering
The Ohio State University
Columbus, USA
bai.426@osu.edu

Alper Yilmaz
Computer Vision Lab.
The Ohio State University
Columbus, USA
yilmaz.15@osu.edu

Halil Sezen
Department of Civil,
Environmental and Geodetic
Engineering
The Ohio State University
Columbus, USA
sezen.1@osu.edu

*Abstract*—Robust Mask R-CNN (Mask Regional Convolutional Neural Network) methods are proposed and tested for automatic detection of cracks on structures or their components that may be damaged during extreme events, such as earthquakes. We curated a new dataset with 2,021 labeled images for training and validation and aimed to find end-to-end deep neural networks for crack detection in the field. With data augmentation and parameters fine-tuning, Path Aggregation Network (PANet) with spatial attention mechanisms and High-resolution Network (HRNet) are introduced into Mask R-CNNs. The tests on three public datasets with low- or high-resolution images demonstrate that the proposed methods can achieve a big improvement over alternative networks, so the proposed method may be sufficient for crack detection for a variety of scales in real applications.

## I. INTRODUCTION

### A. Introduction

In an Artificial Intelligence (AI) application, agents learn from the environment and take actions. Commonly these agents on special tasks use state-of-the-art deep learning methods for solving real world problems. For Structural Damage Detection (SDD) and Structural Health Monitoring (SHM) applications, automation has been attracting attention since the advanced sensors, vision- or vibration-based, are supported with deep learning methods [1]. Thus, platforms like UAVs (Unmanned Aerial Vehicles) and UGVs (Unmanned Ground Vehicles) are employed in field inspections for SDD when human experts cannot timely and safely access to the damaged infrastructures after large wind or earthquake events. Surveillance cameras are installed to monitor the performance of structural members of critical structures, such as important bridges and buildings, during their service life. Depending on its capability to process the information from sensors, an agent for these tasks can be smarter if it interacts with the environment around it [2]. In typical collected images, there are too many items together or cluttered. Some important indicators of structural damage like cracks are usually small and may not be easily noticeable. These indicators may be salient and clear in other special scenarios. Therefore, it is necessary to define the fundamental problem, i.e., scene levels, for this task. It is possible to find robust and accurate models that can be adapted to changing environments if those models can be tested.

When automatic damage detection is used with vision-based technology, there are two main branches: image classification and image segmentation. The goal of classification is to identify the categories of structural attributes such as material type (e.g., steel, concrete, masonry) or structural damage type (e.g., cracking, spalling) without identifying the position of damage in images. On the other hand, image segmentation can detect and mark specific objects, which can be used for quantification in the next step. Image segmentation process labels classes of the damage (semantic segmentation) or partitions individual damage with masks (instance segmentation) for each pixel within the images [3], mainly to detect and delineate different material failure including cracks, spalling and other indicators of structural damage in SDD. In this paper, we focus on crack detection.

Cracking is an initial signal of material failure or even potential structural collapse, but cracks may look totally different in images taken from different angles and distance. Similarly, other types of structural damage manifest themselves differently within complete different scenes. Scene levels of structural damage can be defined as pixel level, object level and structural level [4]. As explained by Bai et al. [5], typically structural components like columns, beams and walls are zoomed in and partially captured at pixel level, but appear complete in object-level images and can be recognized. Meanwhile, an entire building or bridge can be seen in structural-level images. Therefore, typical cracks in pixel-level images are short and wide, while they look long and narrow in object-level images. Cracks also show up in structural-level images whereas they become less visible and are accompanied with crack-like objects as scale increases (see Figure 1). Due to these characteristics, typical images taken at different scales are selected for labeling and are used for training the deep learning models to locate the cracks automatically [5].

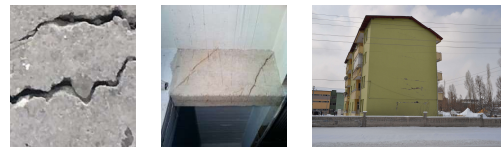

(a) pixel level     (b) object level     (c) structural level
Fig. 1. Three scene levels (scales) in our methods

### B. Problem Statement

The image scale problem in SDD is challenging in current research. For example, the deep learning models in [6, 7, 8]

have a very high accuracy for pixel-level crack detection, however, no details are available about how they may work on larger scales, e.g. at object or structural level. Hoskere et al. [9] apply Residual Network (ResNet) and Visual Geometry Group (VGG) network to identify and segment seven classes of structural damage including cracks. But the accuracy of mask prediction on large-scale images is far from being adequate since only several hundred images are used for training.

We notice the importance of scene levels in SDD and SHM. And we proposed a two-step method, called cascaded network, with a ResNet to classify damage first and a U-Net to finally mark the damage [5]. This is an indirect and time-consuming way and the precision is not high. Therefore, in this paper, we use thousands of images with cracks at various scene levels for training and test state-of-the-art segmentation networks with data augmentation and image enhancement skills. Our tests are based on three public datasets. Thus, we hope to find out how close our proposed models are to field inspections on buildings and bridges damaged by major earthquakes or hurricanes.

## II. RELATED WORK

### A. Deep Learning with R-CNN in Image Segmentation

Major deep learning methods for image segmentation include Fully Convolutional Networks (FCNs), encoder-decoder based models, multi-scale and pyramid network based models, Regional Convolutional Neural Networks (R-CNN) based models, etc.. Each of these techniques has its own contribution and unique aspects, and some of them are typically used as benchmarks in computer vision [3].

In FCNs, multiple convolutional layers are employed on the image directly as the feature extractors with the downsampling and upsampling process inside sliding windows, but its efficiency is very low. With R-CNN, the image is preprocessed and thousands of Region of Interests (RoIs) for feature extraction with FCNs are eventually produced. The R-CNN reduces the computational time compared to alternative approaches and improves the accuracy of segmentation. However, it still is computationally demanding. Fast R-CNN and Faster R-CNN are quite different from the conventional R-CNN. The former applies FCNs directly on the RoIs of the feature maps which comes after convolutional process on the original image, but, in Faster R-CNN, a network called Regional Proposal Network (RPN) on these maps is inserted to automatically produce the proposal, thus the speed and accuracy of prediction are improved [10]. Neither of them are applicable for instance segmentation. He et al. [11] propose a benchmark network, Mask R-CNN, to predict the instance as well as its bounding box and class. There are several significant variations on Mask R-CNN while researchers continue to provide new backbones for feature extraction in image processing. More details about these methods will be discussed when our methods are introduced below.

### B. Structural Damage Detection and Crack Detection

Researchers have already adopted deep learning methods for detecting structural damage. Yeum et al. [12] use AlexNet to classify and identify the structural damages in post-event buildings with large scale images, Hoskere et al. [9] illustrate an experiment with 23-layer ResNet and 9-layer VGG networks to classify and segment seven classes of structural damage, including cracks, spalling, exposed reinforcement, corrosion, fatigue cracks, asphalt cracks, and no damage. Ali et al. [13] introduce Faster R-CNN into defects detection in historical masonry buildings with high-resolution images. Kong and Li [14] describe an application that detects and tracks the propagation of cracks in a steel girder with a video stream. Atha et al. [15] explain the different effects when two algorithms of CNNs are used in detecting metallic corrosion. Gao and Mosalam [16] started the Phi-Net Challenge for collecting pictures of building structural failures in 2018 [16]. Their large dataset, which is used in this paper, is suitable for training and testing different methods for structural damage detection at different scales.

Crack detection with deep learning methods is an active area of research. Zhang et al. [17] propose an improved CNN for autonomous detection of pavement cracks at the pixel level. Liu et al. [8] demonstrate the application with U-Net to segment the crack on concrete structures. Their experiment shows that the proposed network outperform the CNN which was used by Cha et al. [18]. Dung and Anh [19] also use FCNs for localizing the cracks on the concrete surface, Liu et al. [20] implement DeepCrack, which is made of an extended FCN and a Deeply-Supervised Nets (DSN), to pin the pixel-wise cracks. However, these methods of crack segmentation are based on pixel level images and less useful for structural engineering applications.

Recent research in image segmentation have significantly advanced deep learning based SDD and crack detection. Yang et al. [2] employed a hybrid network, composed by Holistically-Nested Edge Detection (HED) network and U-Net, to detect cracks and spalling on concrete structures, and then to reconstruct 3D model via Simultaneous Localization and Mapping (SLAM) for UAV images. Cha et al. [6] applied Fast R-CNN on detecting five types of structural damages, including concrete cracks, steel corrosion with two levels (medium and high), bolt corrosion, and steel delamination. For this purpose, 2,366 images with the size of 500 × 375 were labeled for training. Attard et al. [21] trained a Mask R-CNN with 200 images to locate cracks on the concrete surface at pixel level. Kim and Cho [22] used 376 images in their training data for Mask R-CNN to find the cracks on a concrete wall with high-resolution cameras and utilized an additional image processing procedure on each bounding box to quantitatively measure the width of these cracks. Kalfarisi et al. [23] introduced structured random forest edge detection into bounding boxes of a Faster R-CNN to detect cracks on infrastructures and compared with the performance of Mask R-CNN. A total of 1,250 images were included in training and validation process with the size varying from 344 × 296 to 1,024 × 796. These models are verified with images acquired from field inspections on surface of structural and nonstructural members, including building walls, bridge columns, tunnel walls and roads. The results show that both approaches are robust for this task. Finally, they used photogrammetry software to construct a 3D reality mesh model so that the cracks can be visualized and quantified further [23].

### C. Cascaded Networks for Crack Detection

Some researchers prefer to classify structural damage in SDD because it doesn't require additional resources for labeling the damage. As a result, the location and position of the damage on structural components or structures are unknown until human experts manually check and mark them out. These models can be trained with large number of images while segmentation networks are facing insufficient training samples. In order to inherit the advantage of classification networks, it is necessary to employ a segmentation network to locate the damage after various structural damage have been classified. A cascaded network, including a ResNet and a U-Net, was proposed to perform the detection of cracks [5]. The 152-layer ResNet can meet the requirements of the identification of scene levels, material types, and damage types. Even the severity of structural damage can be classified [24]. U-Net is utilized in the second step to mark the damage like cracks at various scene levels (scales). Then two public datasets are tested by the proposed method. The results show that the cascaded network can improve the accuracy of the detection dramatically in larger scale tasks instead of being limited to pixel-level detection [5]. A comparison between the U-Net of this method and our proposed end-to-end Mask R-CNNs is illustrated in the experiment section of this paper.

## III. FRAMEWORK OF THE PROPOSED METHOD

### A. Data Preparation

We curated a dataset similar to Common Objects in Context (COCO) and used it for training the pipeline because of the fact that the COCO dataset does not contain any structural damage and there are only a few open source datasets available for cracking segmentation. The images selected in our dataset are at various scales, and the tool referred to as the COCO Annotator [25] is used to label cracks for training. Some examples from this process are shown in Figure 2. In these labeled images, cracks are in yellow and background is in purple. Size of the training and labeling images is varied from 168×300 to 4600×3070. By excluding steel structures, 2,021 images are labeled when surface cracks appeared on structural or nonstructural materials at various scales.

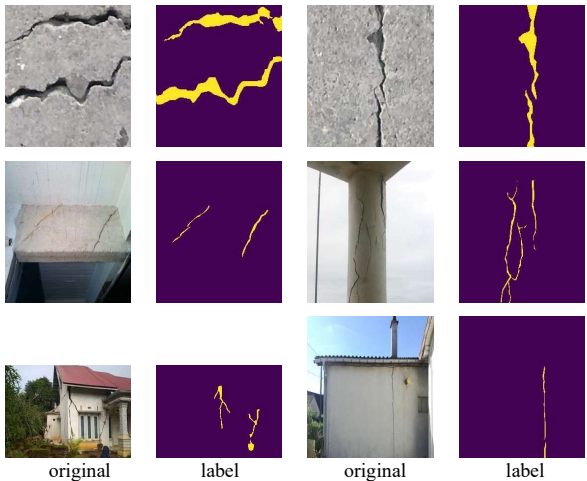

original    label    original    label
Fig. 2. Some examples of training samples in our methods

### B. Data Augmentation with Albumentation

It is a common issue that training data are not sufficiently large for the image segmentation task, so it is necessary to increase the size of training data through augmentation. In this work, we employed Albumentation to augment our training data. Buslaev et al. [26] develop this method with pixel-level transformation and spatial-level transformation, including flipping, rotating, cropping, etc.. Spatial-level transformation is adopted to preprocess our training data since it can change the input images, masks and bounding boxes simultaneously.

### C. Mask R-CNN with Path Aggregation Network (PANet) and Spatial Attention Mechanisms (Mask R-CNN + A-PANet)

He et al. [11] proposed Mask R-CNN for instance segmentation, which is an extension of Faster R-CNN. A RPN is inserted on feature maps to automatically produce RoIs, then a small FCN is applied on each RoI to segment the instance of objects with masks when the classes and bounding boxes of these objects are predicted with the same pipeline as used in Faster R-CNN. In addition, different depth of ResNet and the Feature Pyramid Network (FPN) are combined to extract high-quality feature maps. Since Mask R-CNN is a benchmark for instance segmentation in image processing, many improvements have been made after it was published in 2017. The framework of Mask R-CNN is shown in Figure 3. Liu et al. [27] improved Mask R-CNN by replacing FPN with PANet to gain better performance. Because features of low layers in the pyramid can reach high layers by skip-connections and a technique called adaptive feature pooling can fuse all levels of features for each proposal, their proposed method achieves a higher accuracy when a modified approach on mask prediction is also adjusted. Figure 4 shows the framework of PANet, which is part of methodologies used in this paper. Furthermore, inspired by Nie et al. [28] on their application to detect ships from satellite images, we also introduce spatial attention mechanisms as suggested by Zhu et al. [29] into our methods. The goal of this study is to facilitate the backbone of Mask R-CNN to extract more useful features in crack detection. We call this method as Mask R-CNN + A-PANet in this paper.

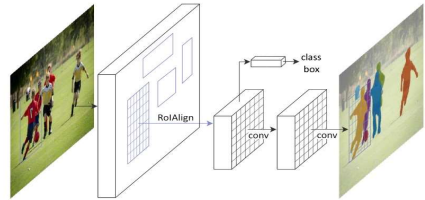

Fig. 3. The Mask R-CNN framework for instance segmentation [11]

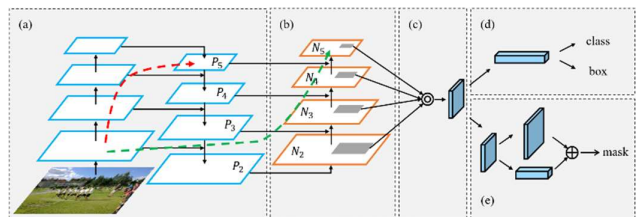

Fig. 4. Illustration of PANet framework. (a) FPN backbone. (b) Bottom-up path augmentation. (c) Adaptive feature pooling. (d) Box branch. (e) Fully connected fusion. Note that we omit channel dimension of feature maps in (a) and (b) for brevity [27].

### D. Mask R-CNN with High-resolution Network (Mask R-CNN + HRNet)

CNNs are backbones of segmentation networks. For example, the original of Mask R-CNN uses a 101-layer ResNet as its backbone. Sun et al. [30] developed a new network named HRNet to extract features from an original image. Utilizing repeated multi-scale fusions across these convolutional blocks, this network maintains high-resolution representations via inter-connection between high- and low-resolution convolutional modules within a parallel structure. As shown in Figure 5, there are four stages in HRNet. High-resolution features are kept until the end of convolutional

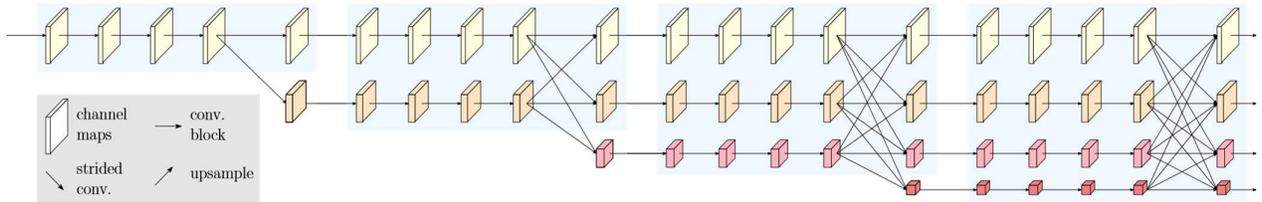
Fig. 5. A simple example of a high-resolution network. There are four stages. The first stage consists of high-resolution convolutions. The second (third, fourth) stage repeats two-resolution (three-resolution, four-resolution) blocks [30].

operation, and low-resolution ones are added to each new stage. The connection between them may be the key for better feature extraction. In this paper, HRNet is also employed for crack detection as the backbone of a Mask R-CNN and named as Mask R-CNN + HRNet.

## IV. IMPLEMENTATION

Our implementation with these Mask R-CNNs are originated from MMDetection source codes [31]. Modifications are made to the training and testing process for consideration of our specific problem and data. First, these augmented images are trained and evaluated. Then, the data from Phi-Net [16], the 2017 Pohang earthquake images [32] and the 2017 Mexico City earthquake images [33] at various scales are tested. The hyperparameters are defined as follows: learning rate is 0.002, momentum is 0.9 and decay rate of weights is 0.0001. The loss function for mask is "cross-entropy" and for bounding boxes is "smooth L1". The training and testing for the model are using NVIDIA GeForce GTX 2080 Super. The total number of epochs for each training model is 100.

### A. Evaluation on the proposed models

Because the labeled data in training and validation have the same format as COCO data, we followed the same standard metrics to evaluate our models via assessing them with our validation dataset. The result is shown in Table I. The AP (Average Precision) is based on Intersection over Union (IoU), while $AP$, $AP_{50}$, $AP_{75}$, and $AP_S$, $AP_M$, $AP_L$ are defined with different threshold values and scales. Here we report Mask AP of box as most researchers have done. Both of the original Mask R-CNN and the Mask R-CNN + A-PANet employ 50-layer ResNet as the backbone, and the Mask R-CNN + HRNet uses four stage high-resolution networks. The AP metrics of our proposed Mask R-CNNs are improved dramatically. This means the latest development of Mask R-CNN has increased the chance to solve real crack detection problem.

TABLE I
COMPARISON OF MASK R-CNNS WITH VALIDATION DATA

| Methods | AP | $AP_{50}$ | $AP_{75}$ | $AP_S$ | $AP_M$ | $AP_L$ |
|---|---|---|---|---|---|---|
| Mask R-CNN | 21.7 | 54.9 | 16.6 | 28.6 | 41.2 | 23.1 |
| Mask R-CNN + A-PANet | 46.9 | 78.5 | 48.9 | 70.0 | 53.9 | 41.5 |
| Mask R-CNN + HRNet | **59.3** | **86.7** | **63.6** | **80.0** | **58.4** | **62.2** |

The criterion for a valid prediction in the following tests is defined as at least one crack being located by a bounding box or a mask.

### B. Testing on Phi-Net datasets [16]

The Phi-Net datasets, in which the images have been rescaled to 224 × 224, are tested. There are three kinds of datasets, pixel-level, object-level and structural-level images with or without cracks. We also compared with the prediction by U-Net in the cascaded network [5]. In the predictions of the proposed approaches, Mask R-CNN + A-PANet is the only method being used here because it works better in Phi-Net datasets based on our observation. After considering the effect of low resolution in the Phi-Net datasets, the threshold for these cracks being detected is set as 0.2 instead of 0.5 as most studies used.

The metrics including recall, precision and total accuracy are used here:

$$Recall = \frac{TP}{TP+} \quad (1)$$

$$Precision = \frac{TP}{TP + FP} \quad (2)$$

$$Accuracy = \frac{TP+TN}{TP + FP + FN + FN} \quad (3)$$

where *TP* and *TN* are true positive and negative, *FP* and *FN* are false positive and negative, respectively.

*1) Pixel scene level Task in Phi-Net:* There are 4,663 images in this dataset. The accuracy of prediction from U-Net is 60.5% as an end-to-end network for crack detection. With Mask R-CNN + A-PANet, the accuracy and the recall of prediction can reach to 84.7% and 77.4% respectively for detecting and locating the actual cracks.

Figure 6 shows some examples of correct prediction from U-Net and Mask R-CNN + A-PANet. Cracks are marked with red color from U-Net model in the middle images. In the images on the right, the Mask R-CNN gives the masks and bounding boxes in purple and green colors, respectively. These colors have the same meaning in Figures 7, 8 and 9.

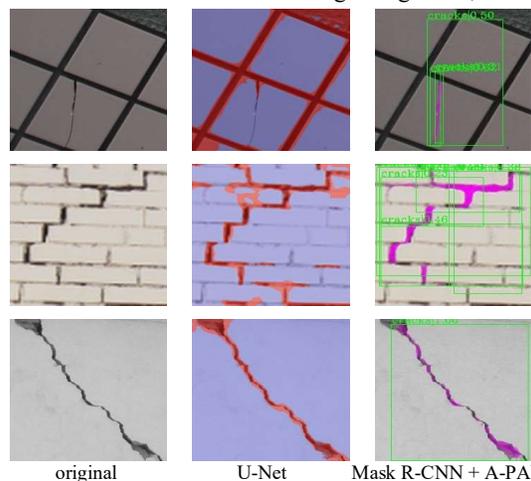

original      U-Net      Mask R-CNN + A-PANet
Fig. 6. Some correct predictions of U-Net and Mask R-CNN + A-PANet at pixel level.

*2) Object scene level Task in Phi-Net:* We tested 5,713 images in this dataset. The accuracy of the prediction from U-Net is 26.2% as an end-to-end network for crack detection, but Mask R-CNN + A-PANet can obtain an accuracy of 77.1%. Its recall can achieve 62.3% while it is 59.6% in U-Net model. It should be noted that at this scale there are more objects like wires, windows and doors in images. Figure 7 shows good predictions of both U-Net and Mask R-CNN + A-PANet.

*3) Structural scene level Task in Phi-Net*: A total of 5,832 images are included in this dataset. The accuracy of the U-Net model is 8.9% whereas Mask R-CNN + A-PANet can reach to 81.9% accuracy. The recall of the later is 40.9% while that for the U-Net is 49.7%. It should be noted that in this scale more objects (like cables, tree branches, and other objects) are shown up in these images when cracks are less visible. On the other hand, because there are so many images without any cracks, and there are cracks less visible at such a large scale with such low resolution, the overall accuracy of Mask R-CNN + A-PANet can be very high as the recall and precision can be low. However, it should be pointed out that the predictions of crack location by Mask R-CNN + A-PANet is more accurate than those from the U-Net (see Figure 8).

*4) Direct test on Task 8 in Phi-Net:* We tested the Mask R-CNN model at various scene levels in the previous step. We also wanted to know its performance by testing images at different scales that are mixed together, which is more common in field inspections. In Task 8 of the Phi-Net [16], there are four crack types, including non-cracking, flexural cracks, shear cracks and combined cracks. We incorporated all these images with different cracks into a cracking class while keeping the noncracking ones as another class. A total of 1,502 images for non-cracking and 1,130 images for cracking cases were tested with Mask R-CNN + A-PANet and U-Net.

The results in Table II indicate that U-Net can lead to a higher recall than Mask R-CNN + A-PANet for this dataset. The table also shows that U-Net cannot discern those images without any cracks as well as in the previous tests. Thus, the accuracy and precision from U-Net are much lower than Mask R-CNN + A-PANet. However, failure examples in Figure 9 show that the objects like wires, edges of windows, doors and tiles, and trees are the main distractions being detected by both models.

TABLE II
PREDICTIONS OF U-NET AND R-CNN + A-PANET ON PHI-NET TASK 8

| Methods | Accuracy | Recall | Precision |
|---|---|---|---|
| U-Net | 44.7% | **98.5%** | 43.6% |
| Mask R-CNN + A-PANet | **75.1%** | 67.7% | **82.4%** |

Figure 9 indicates that the Mask R-CNN model still has difficulties to handle these crack-like distractions, even the shadows can cause erroneous prediction. In addition, all these testing images are in low resolution while their size is uniformly set as 224 × 224. This size may be good enough for detecting pixel-level cracks and part of object-level cracks, but not at the rest levels. This is one of the main reasons for the poor performance on detecting cracks at larger scales.

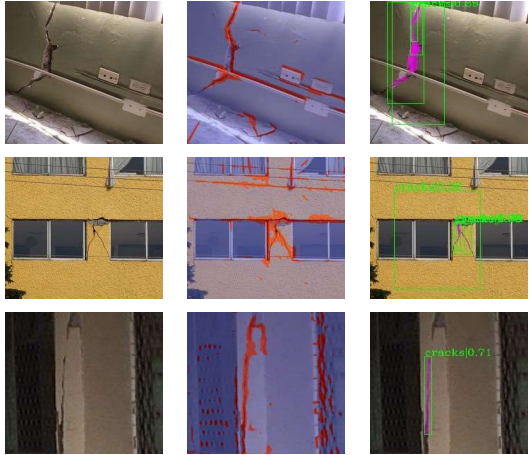
      original           U-Net      Mask R-CNN + A-PANet
Fig. 7. Some correct predictions of U-Net and Mask R-CNN + A-PANet at object level.

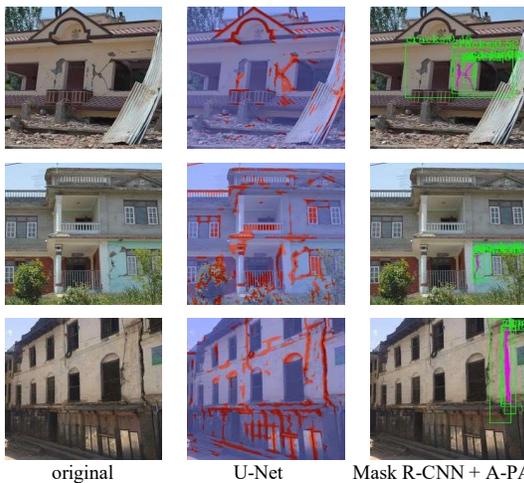
      original           U-Net      Mask R-CNN + A-PANet
Fig. 8. Some correct predictions of U-Net and Mask R-CNN + A-PANet at structural level

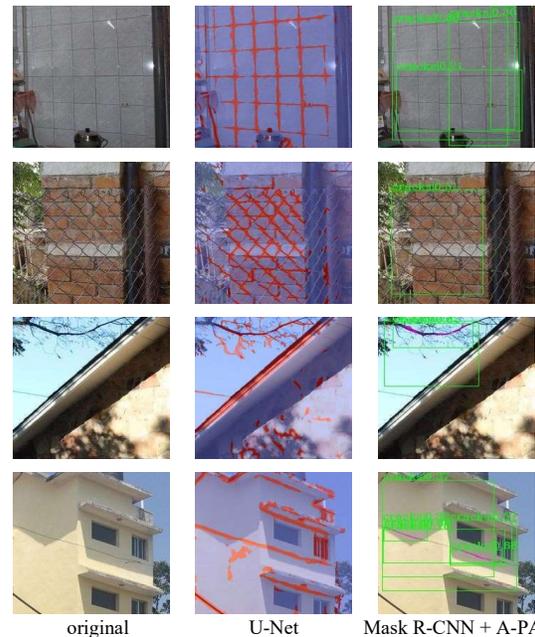
      original           U-Net      Mask R-CNN + A-PANet
Fig. 9. Some incorrect predictions of U-Net and Mask R-CNN for Phi-Net data

## C. Testing on the Other Two Datasets

In field inspections, high-definition cameras are commonly used. We believe that cracks can be more easily detected with high-definition images because the more pixels, the clearer appearance of cracks. We began with a dataset of our own, in which the image size is varied from 1600 × 1200 to 4290 × 3200, so that the parameters of these Mask R-CNNs could be fine-tuned. With this, it is possible to predict and mark more cracks in the images. The following parameters are used for 2017 Pohang earthquake images [32] and 2017 Mexico City earthquakes images [33]: 1) the threshold for the cracks being detected is set to be 0.5 as the original Mask R-CNN does, and 2) the threshold for mask is reduced from 0.5 to 0.05 for inference. A comparison between Mask R-CNN with A-PANet and HRNet is done here while the same detection criterion is used as before.

*1) Testing on 2017 Mexico City earthquake images [33]:* There are 4,136 images with two image sizes, 2740 × 3650 and 6000 × 4000, in this dataset. All of them are taken by human experts at Purdue University when they conducted the field investigation in Mexico City after a magnitude 7.1 earthquake in 2017. Figure 10 shows some examples of the prediction from our models. The cracks are marked in purple while the bounding boxes are in green in both Mask R-CNN models. Colors have the same meaning in Figures 10 and 11.

The accuracy, recall and precision of the two models are shown in Table III, two of them are very close except recall for which there is a 9.1% difference.

TABLE III
PREDICTIONS OF MASK R-CNN ON 2017 MEXICO CITY EARTHQUAKE

| Methods | Accuracy | Recall | Precision |
|---|---|---|---|
| Mask R-CNN + A-PANet | 70.6% | 53.6% | **92.9%** |
| Mask R-CNN + HRNet | **73.0%** | **62.7%** | 90.5% |

*2) Testing on 2017 Pohang earthquake images [32]:* In this dataset, a team of researchers supported by the American Concrete Institute (ACI) collected images during their field inspection after a magnitude 5.4 earthquake in Pohang, South Korea, in 2017. The total number of images used for testing is 4,109. The images in this dataset have two sizes, 2600 × 3890 and 5180 × 3460. Some examples of the prediction are shown in Figure 11. Accuracy, recall and precision of the two models are shown in Table IV.

TABLE IV
PREDICTIONS OF MASK R-CNN ON 2017 POHANG EARTHQUAKE

| Methods | Accuracy | Recall | Precision |
|---|---|---|---|
| Mask R-CNN + A-PANet | **74.1%** | 56.9% | **94.7%** |
| Mask R-CNN + HRNet | 74.0% | **63.6%** | 88.3% |

Table III and IV show that precision for Mask R-CNN + A-PANet is higher than Mask R-CNN + HRNet while recall of the former is lower than the later. But their accuracy is close.

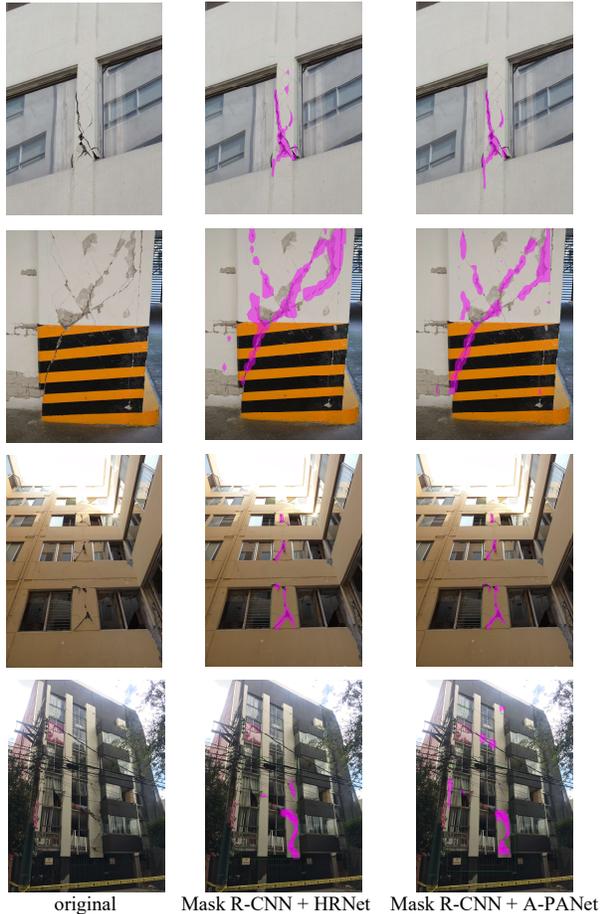

original    Mask R-CNN + HRNet    Mask R-CNN + A-PANet
Fig. 10. Prediction of Mask R-CNN with Attention PANet and HRNet for 2017 Mexico City earthquake images [33].

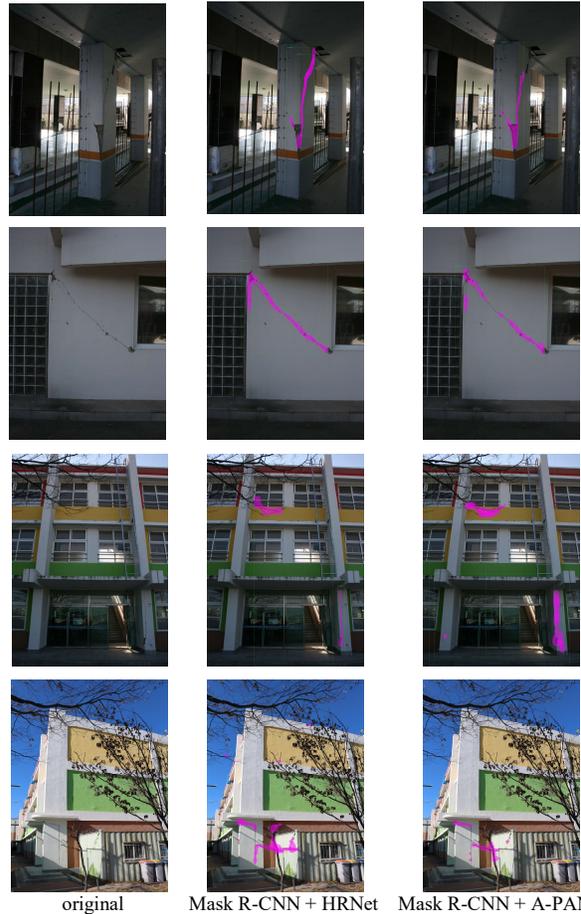

original    Mask R-CNN + HRNet    Mask R-CNN + A-PANet
Fig. 11. Prediction of Mask R-CNN with Attention PANet and HRNet for 2017 Pohang earthquake images [32].

*3) Comments on the Testing Results of Two Datasets:* Tests show that cracks are more visible and easier to be

detected in high-definition images by our models. The proposed end-to-end networks can reach a high accuracy for detecting cracks on infrastructures after a disaster including large earthquakes. Furthermore, it was also discovered in this research that an appropriate angle and a right distance to the observed objects are critical because they affect detection rate even with high quality images. However, crack-like things also distract the prediction of the models, which leads to inaccurate detection. As a result, the proposed two Mask R-CNNs work well at pixel level and object level but get worse at larger scale.

## V. Discussion

In this work, the scene levels or scales are treated as a fundamental problem in detecting small damage like cracks in real world. So the models proposed in this paper are evaluated using different publicly available datasets with an aim to find end-to-end networks to segment them automatically and accurately, regardless of its number, shapes and positions. Some parameters used in the original Mask R-CNN are identified to be fine-tuned in order to robustly detect the damage in 2D images.

- With the latest developments in deep learning, end-to-end networks make it possible to automatically recognized damage from images of field inspections after a large earthquake or a hurricane. Through tests conducted on various selected images, we observed that these proposed models can give very precise prediction for instance of pixel-level and object-level cracks, but they don't work well at the structural level. In addition to insufficient number of samples at such a large scale included in our training data, thus may be a result of distraction due to trees, wires, and other crack-like things in the images. However, high-resolution images are helpful for improving the models, especially with HRNet, to detect real cracks. This is in line with our goal to lower the effect of scale changing in structural damage detection.

- Low resolution is commonly used with high speed cameras whereas high resolution is standard for other high quality data collection. Our testing results may be useful for other researchers when they try to balance the speed and resolution in their field inspection missions. For example, Mask R-CNN + A-PANet can be employed in video detection task and Mask R-CNN + HRNet is applicable in few shot task. We are currently working on consistency prediction between frames with these models.

- The inference speed for high-definition images is much slower than for low-definition ones in the two Mask R-CNNs. This is because more pixel-wise processes are involved that leads to more time consuming. In addition, the masks from the models do not exactly fit the shapes and positions of cracks in some cases, and not every crack is marked separately. Therefore, we are still in the process of exploring potential solutions for this problem.

- Better detection rate can be achieved only if the appropriate distance and viewpoints are available in the images from field investigations. Therefore, it is feasible to train the agents to interact with the environment to gain higher quality data in this task.

## VI. Conclusions

In this paper, 2,021 images are selected and labeled as training and validation data at various scales for structural damage detection on surface cracks. The study is intended to find robust end-to-end instance segmentation networks for crack detection of visual data in field inspections after extreme events such as large earthquakes and hurricanes. Several public datasets with earthquake-induced damage are tested by our proposed Mask R-CNNs with HRNet as the backbone and with PANet combining spatial attention mechanism. Our tests show that the proposed models are applicable for real practice and robust to overcome the influence of scale difference. Thus, it is possible that agents such as UGVs and UAVs are smart enough to automatically detect the indicators of material failure or structural collapse, especially by interacting with the environment around them. In addition, we also show how to adjust parameters to address our concerns that cracks are identified and marked as accurately as possible.

Our future work is to generate more labeled data, in particular larger number of large-scale images, for training purposes. We also plan to add more validation and testing data for real application. There is a great potential for other deep learning methods on instance segmentation. Moreover, the speed and accuracy of structural damage detection need to be improved to automatically identify damage from images collected during field inspections.